\documentclass[sigconf, nonacm]{acmart}
\AtBeginDocument{%
  }

\setcopyright{acmlicensed}
\copyrightyear{2018}
\acmYear{2018}
\acmDOI{XXXXXXX.XXXXXXX}
\acmConference[Conference acronym 'XX]{Make sure to enter the correct
  conference title from your rights confirmation email}{June 03--05,
  2018}{Woodstock, NY}
\acmISBN{978-1-4503-XXXX-X/2018/06}

\usepackage{booktabs}
\usepackage{multirow}
\usepackage{xcolor}
\usepackage{adjustbox}
\usepackage[many]{tcolorbox}

\newtcolorbox{promptbox}[1]{
  colback=gray!5, colframe=gray!60!black, fonttitle=\bfseries\small,
  title={#1}, boxrule=0.5pt, arc=2pt,
  left=6pt, right=6pt, top=4pt, bottom=4pt,
  fontupper=\small
}

\newcommand{\bench}{IKEA-Bench}
\newcommand{\pp}{\,pp}

\newcommand{\eg}{\textit{e.g.}}

\newcommand{\etal}{\textit{et al.}}

\begin{document}

\title{Benchmarking and Mechanistic Analysis of Vision-Language Models for Cross-Depiction Assembly Instruction Alignment}

\author{Zhuchenyang Liu}
\affiliation{%
  \institution{Aalto University}
  \city{Espoo}
  \country{Finland}}
\email{zhuchenyang.liu@aalto.fi}

\author{Yao Zhang}
\affiliation{%
  \institution{Aalto University}
  \city{Espoo}
  \country{Finland}}
\email{yao.1.zhang@aalto.fi}

\author{Yu Xiao}
\affiliation{%
  \institution{Aalto University}
  \city{Espoo}
  \country{Finland}}
\email{yu.xiao@aalto.fi}

\begin{abstract}


2D assembly diagrams are often abstract and hard to follow, creating a need for intelligent assistants that can monitor progress, detect errors, and provide step-by-step guidance. In mixed reality settings, such systems must recognize completed and ongoing steps from the camera feed and align them with the diagram instructions. Vision Language Models (VLMs) show promise for this task, but face a depiction gap because assembly diagrams and video frames share few visual features. To systematically assess this gap, we construct \bench{}, a benchmark of 1,623 questions across 6 task types on 29 IKEA furniture products, and evaluate 19 VLMs (2B--38B) under three alignment strategies. Our key findings: (1) assembly instruction understanding is recoverable via text, but text simultaneously degrades diagram-to-video alignment; (2) architecture family predicts alignment accuracy more strongly than parameter count; (3) video understanding remains a hard bottleneck unaffected by strategy. A three-level mechanistic analysis further reveals that diagrams and video occupy disjoint ViT subspaces, and that adding text shifts models from visual to text-driven reasoning. These results identify visual encoding as the primary target for improving cross-depiction robustness. Project page: \url{https://ryenhails.github.io/IKEA-Bench/}


\end{abstract}



\begin{teaserfigure}
  \includegraphics[width=0.95\textwidth]{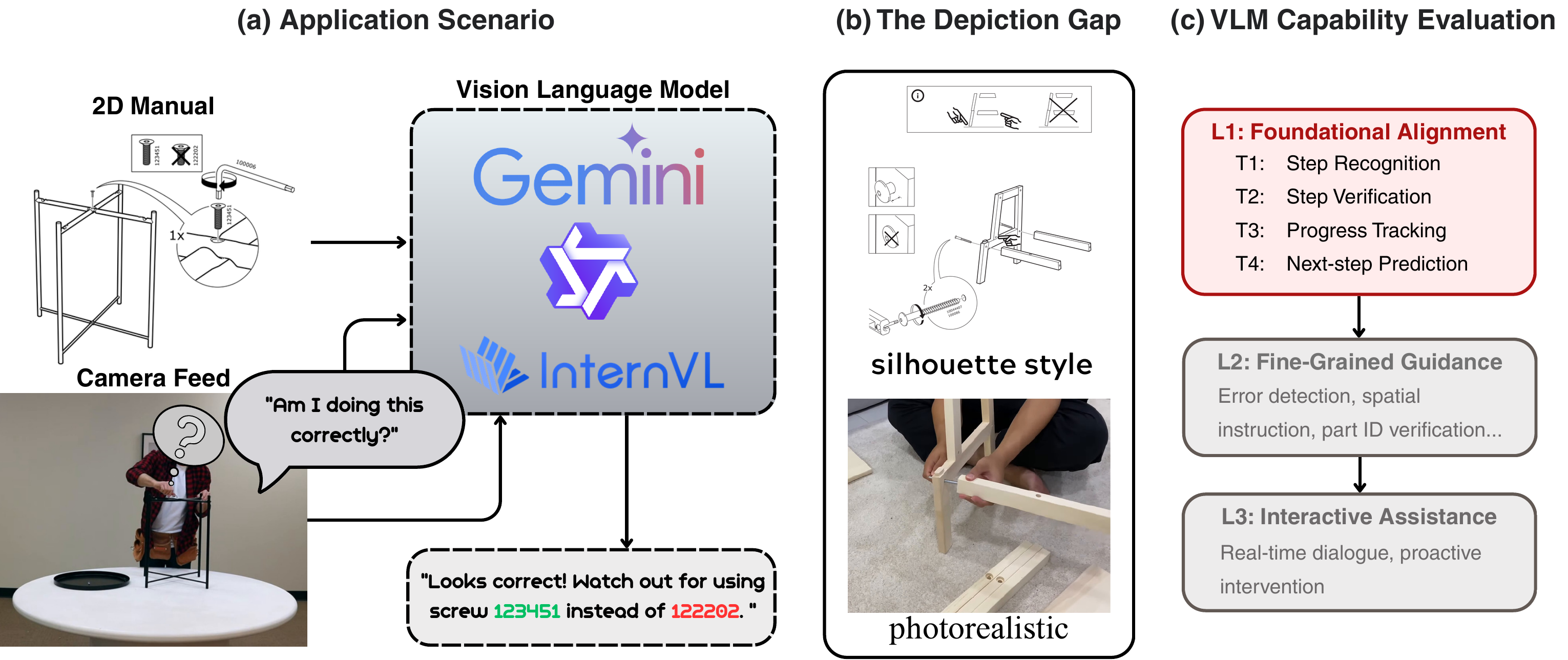}
  \caption{(a) Illustration of 2D-manual-based assembly guidance that leverages a VLM to align schematic diagrams with real-time camera feed. (b) The depiction gap: diagrams and video frames share no visual features despite depicting the same step. (c) The foundational alignment level (L1)  remains unsolved.}
  \Description{Three panels. Left: application scenario showing a 2D manual and camera feed input to a VLM. Center: the depiction gap between silhouette-style diagrams and photorealistic video. Right: three-level capability hierarchy with L1 highlighted and L2-L3 grayed out.}
  \label{fig:teaser}
\end{teaserfigure}

\received{20 February 2007}
\received[revised]{12 March 2009}
\received[accepted]{5 June 2009}

\maketitle


\section{Introduction}

In everyday life, many products---especially furniture---arrive as flat packs that customers assemble using 2D diagrams. For example, IKEA alone provides wordless, diagram-based manuals for more than 9,500 products. These diagrams are often abstract and hard to follow, highlighting the need for intelligent digital assistants that can monitor progress, detect errors, and provide step-by-step guidance. Emerging mixed-reality assembly assistants address this need by detecting the user’s current step from a live video feed, matching it to the corresponding instruction, and overlaying guidance onto the real-world view~\cite{Wang2023HoloAssistAE, Sener2022Assembly101AL}. Technically, it requires alignment between instructions extracted from diagrams and video frames.


Vision-language models (VLMs) offer a scalable alternative: rather than requiring 3D assets, a VLM could read a product’s 2D assembly diagram and align its instructions with a camera feed zero-shot (Figure~\ref{fig:teaser}a). Recent work has begun exploring VLMs for assembly-related tasks, though not yet for diagram-to-video alignment. Manual2Skill~\cite{Tie2025Manual2SkillLT} uses a VLM to parse manual images into structured assembly graphs for robotic manipulation, but reports only 62\% success rate on graph generation (F1\,=\,0.684), with failures largely attributed to inaccurate diagram interpretation. LEGO Co-builder~\cite{Huang2025LEGOCE} benchmarks VLMs on assembly state detection from synthetic images and finds that GPT-4o achieves only 40.5\% F1. These results indicate that VLM understanding of assembly instructions from visual inputs remains limited, yet no prior work has systematically evaluated the diagram-to-video alignment that real-world assembly guidance requires.


What makes diagram-to-video alignment particularly challenging is what we term the \emph{depiction gap} (Figure~\ref{fig:teaser}b). Cross-depiction recognition---the task of recognizing objects across different visual styles such as photographs, paintings, and sketches---is a known challenge in computer vision~\cite{Cai2015TheCP, Hall2015CrossdepictionPR}. Assembly diagrams pose a harder variant: diagrams use arrows, exploded views, and part silhouettes, while video shows hands assembling furniture in cluttered, photorealistic scenes. These two depictions do not share visual features despite depicting the same procedure. Moreover, unlike prior cross-depiction work that studies static object recognition~\cite{Cai2015TheCP, Hall2015CrossdepictionPR, Koley2024HowTH}, the depiction gap involves temporal procedures where the correct match depends on which parts have been assembled so far. To the best of our knowledge, this depiction gap has not been systematically evaluated due to the lack of a benchmark.


To address this issue, we introduce \bench{}, a benchmark for evaluating VLMs on foundational diagram‑to‑video alignment. As illustrated in Figure~\ref{fig:teaser}c, it covers four tasks central to manual‑based assembly guidance: T1 step recognition, T2 step verification, T3 progress tracking, and T4 next‑step prediction. We further include two diagnostic tasks to isolate bottlenecks: D1 (video discrimination) tests whether the model can discriminate between video segments of different steps, and D2 (instruction comprehension) tests whether it can order assembly diagram steps correctly. \bench{} comprises 1,623 questions across these six task types, built from 29 IKEA furniture products in the IKEA Manuals at Work dataset\cite{Liu2024IKEAMA}. The benchmark also supports three alignment strategies—Visual (diagrams only), Visual+Text (diagrams with text descriptions), and Text‑Only (text replacing diagrams), enabling practitioners to identify which input configuration works best for each subtask.


Using \bench{}, we evaluate 19 VLMs—17 open‑source models spanning eight architecture families (2B–38B parameters) and two proprietary models (Gemini 3.1 Pro, Gemini 3 Flash~\cite{geminiteam2025geminifamilyhighlycapable})—across all three alignment strategies. 
This evaluation yields three key findings. First, the ability to comprehend assembly diagram content (tested by D2) is recoverable via text-mediated alignment (+23.6\pp{}), but counter-intuitively, adding text simultaneously degrades diagram-to-video alignment on T1 ($-$3.1\pp{} on average). Second, architecture family predicts cross-depiction alignment accuracy more strongly than parameter count. Third, video discrimination (D1) remains low across all models and all strategies, forming a hard ceiling that no alignment strategy overcomes.

To understand why all evaluated models struggle at cross-depiction alignment, we further conduct a three-layer mechanistic analysis on four representative open-source VLMs. Layer~1 (ViT representations) tests whether the visual encoder creates a shared space for diagrams and video. Layer~2 (LLM hidden states) tests whether adding text reduces the model’s reliance on diagram representations. Layer~3 (attention routing) tests whether text redistributes attention away from visual modalities. We find that diagrams and video occupy disjoint representational subspaces at the ViT level (CKA~$\approx$~0), and that adding text triggers a measurable shift from visual to text-driven reasoning, confirmed at both the representation and attention levels.

In summary, our contributions are:
(1) The first diagram-to-video alignment benchmark for assembly guidance, comprising 1,623 questions across 6 task types and 3 alignment strategies on 29 IKEA products.
(2) A three-layer mechanistic analysis that localizes cross-depiction failure to the visual encoder and reveals that text distracts models from visual processing.
(3) Empirically grounded design guidelines for alignment strategy selection, architecture choice, and directions for future improvement.

\begin{figure*}[tb]
  \centering
  \includegraphics[width=\textwidth]{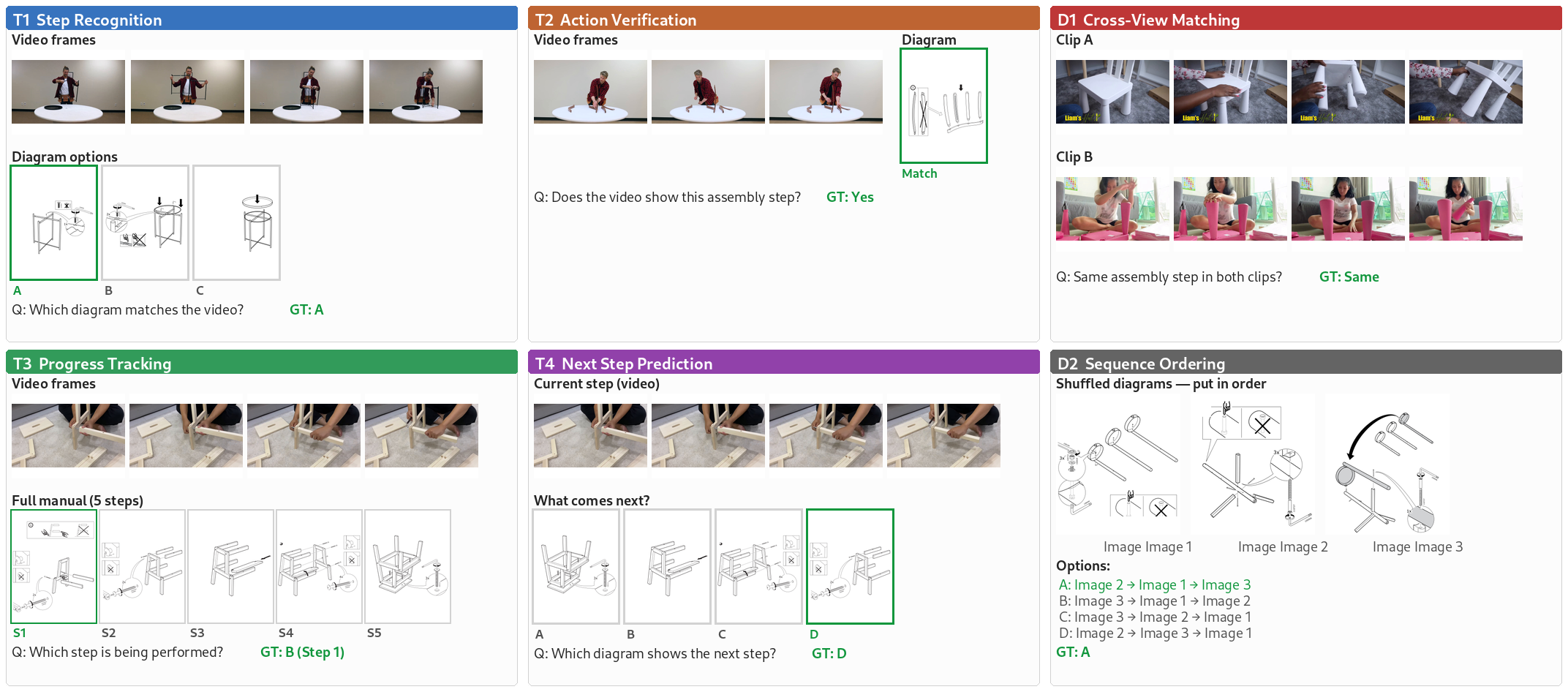}
  \caption{One example per task type. T1--T4 test cross-depiction alignment at increasing difficulty; D1--D2 isolate video and assembly instruction understanding respectively. Green borders mark correct options. Full prompt templates are in Suppl.~A.}
  \Description{Six panels in a 3x2 grid, each showing one task type with video frames, diagram options, question, and correct answer highlighted in green.}
  \label{fig:tasks}
\end{figure*}

\section{Related Work}

Our work lies at the intersection of cross-depiction recognition and VLM evaluation for procedural understanding. We review prior work along these two threads below.

\subsection{Cross-Depiction Recognition and Assembly Understanding}

\paragraph{Cross-depiction recognition.}
Cai~\etal{}~\cite{Cai2015TheCP} formalized the cross-depiction problem as recognizing objects across photographs, paintings, and drawings. Hall~\etal{}~\cite{Hall2015CrossdepictionPR} showed that spatial-part models outperform appearance-based methods across depiction styles, since appearance features tend to overfit to a single depiction. More recently, sketch-based image retrieval has extended this line to handle varying abstraction levels~\cite{Koley2024HowTH}. All prior work studies static object recognition across visual styles. Our work extends cross-depiction recognition to a harder setting: matching between schematic assembly diagrams and photorealistic video within a temporal procedural context.


CaptainCook4D~\cite{Peddi2023CaptainCook4DAD} and SPLICE~\cite{Ballout2025CanYS} evaluate temporal and causal reasoning in procedural video. Shin~\etal{} ~ \cite{Shin2025DoVL} investigate whether video-language models genuinely exploit temporal structure or rely on static frame-level cues. All of these benchmarks use text instructions as the procedural specification. Visual instructions such as assembly diagrams have not been tested as an instruction modality.

\subsection{VLM Evaluation and Analysis}

\paragraph{VLM benchmarks for technical images.}
Recent work has begun probing VLMs on non-photographic visual inputs. Chain-of-Region~\cite{Li2025ChainofregionVL} decomposes complex scientific diagrams into sub-regions for improved reasoning. TechING~\cite{Nadeem2026TechINGTR} benchmarks flowchart and block-diagram understanding. MIMIC~\cite{Das2026MoreIM} reveals that VLMs fail to aggregate information across multiple images. These benchmarks evaluate diagram comprehension or multi-image reasoning in isolation. Our benchmark differs in that it tests diagram-to-video matching within a procedural context.

\paragraph{Mechanistic analysis of VLMs.}
On the analytical side, the Hidden Life of Tokens~\cite{Li2025TheHL} traces how visual information degrades through generation layers, while NOTICE~\cite{Golovanevsky2024WhatDV} identifies universal attention heads that mediate multimodal integration. Our three-layer mechanistic analysis builds on these methods but addresses a different question: rather than studying how visual information is processed within a single depiction, we analyze why VLMs fail to align two fundamentally different depictions of the same content.


\section{IKEA-Bench}

We describe the source data, task design, question construction, and alignment strategies below.

\subsection{Source Data}

Several datasets provide annotations for furniture assembly. IKEA-Manual~\cite{Wang2023IKEAManualSS} pairs 3D furniture models with manual segmentation and assembly plans. Assembly101~\cite{Sener2022Assembly101AL} and IKEA ASM~\cite{BenShabat2020TheIA} provide multi-view video with action and pose annotations. Among these, IKEA Manuals at Work~\cite{Liu2024IKEAMA} is the only dataset that provides dense temporal alignment between manual step diagrams and assembly video segments, making it uniquely suited for constructing diagram-to-video alignment tasks.

\bench{} is built on IKEA Manuals at Work, which covers 36 furniture products with corresponding assembly videos, wordless manual step diagrams, and temporal annotations marking when each step begins and ends in every video. We use 29 of the 36 products; 4 are excluded for having $\leq$2 assembly steps (insufficient for tasks requiring $\geq$3 steps), and 3 for data quality issues (blank or duplicate step images). From the retained products, we extract 2,569 video frames across 97 assembly videos. Questions are deterministically constructed from the temporal annotations (\S\ref{sec:construction}), yielding 1,623 benchmark questions. Full statistics are provided in Suppl.~B.

IKEA manuals are well suited for studying diagram-to-video alignment. They are wordless by design, maximizing the visual-only nature of the assembly instruction. Their fine-grained sequential structure produces natural hard negatives: adjacent steps depict the same parts but in different stages of assembly---for example, step~3 may show a leg positioned next to the seat, while step~4 shows the same leg screwed into place.

\subsection{Task Design}
\label{sec:taskdesign}

An assembly guidance system must be able to recognize the current step, verify correctness, and predict what comes next. We select 6 task types that test these three prerequisite capabilities, plus two diagnostic controls that isolate individual bottlenecks. Figure~\ref{fig:tasks} shows one example per task.

``Which step am I on?'' maps to step recognition (T1: match video to one of 4 diagram options; T3: locate the current step among all diagrams for a product). ``Am I doing this correctly?'' maps to step verification (T2: judge whether a video segment matches a given diagram). ``What comes next?'' maps to next-step prediction (T4: identify the diagram for the step after the one shown in video). Two diagnostic tasks isolate unimodal understanding: D1 (video discrimination) tests whether the model can distinguish two video segments showing different assembly steps; D2 (instruction comprehension) tests whether it can order three consecutive diagram steps correctly. 

\begin{table}[tb]
  \caption{Task summary.}
  \label{tab:tasks}
  \small
  \begin{tabular}{lllrc}
    \toprule
    Task & Input & Format & $N$ & Chance \\
    \midrule
    T1 & V+4P & 4-MC (image) & 320 & 25\% \\
    T2 & V+1P & Binary & 350 & 50\% \\
    T3 & V+all P & 4-MC (text) & 334 & 25\% \\
    T4 & V+4P & 4-MC (image) & 204 & 25\% \\
    D1 & V+V & Binary & 350 & 50\% \\
    D2 & 3P & 4-MC (text) & 65 & 25\% \\
    \bottomrule
  \end{tabular}
  \\[2pt]
  {\footnotesize V = video frames; P = plan (assembly instruction diagram); $n$P = number of diagrams provided; 4-MC = 4-way multiple choice; $N$ = number of questions; Chance = random-guess accuracy.}
\end{table}

Table~\ref{tab:tasks} summarizes the input, format, and scale of the 6 task types. T1 and T4 present answer options as diagram thumbnails rather than text labels, forcing visual parsing of schematic content. T3 includes all manual step diagrams for a product (3--10, median~4). D2 requires 3 consecutive manual step diagrams; only consecutive triplets are used because non-adjacent steps are visually too dissimilar, making ordering trivial via coarse part matching. Across all 29 products, there are exactly 65 valid consecutive triplets, yielding 65 D2 questions.

\paragraph{Anti-shortcut design.}
The core design principle is preventing shortcut strategies. Distractors are drawn from adjacent assembly steps that depict the same parts in different stages of assembly, forcing genuine cross-depiction understanding rather than coarse part-presence heuristics. Binary tasks (T2, D1) randomize option order (ground-truth distributions: T2 53\%/47\% No/Yes; D1 47\%/53\% Different/Same). After randomization, aggregate prediction rates across 17 models are balanced (T2: 49\% A-rate; D1: 52\% A-rate), confirming that the benchmark does not induce systematic prediction bias.
 
\subsection{Question Construction}
\label{sec:construction}
 
All questions are deterministically constructed from ground-truth temporal alignment annotations . For each step, video frames are uniformly sampled from the annotated temporal segment (4--8 frames depending on segment length). Distractors for T1 and T4 are drawn from adjacent assembly steps, ensuring high visual similarity. T2 pairs each video segment with either the matching diagram (positive) or an adjacent-step diagram (negative), balanced 53\%/47\%. T3 presents all manual step diagrams for a product as options with text labels. D1 pairs video segments from the same step (positive) or different steps of the same product (negative). D2 selects 3 consecutive manual diagrams and generates all valid ordering permutations as distractors. Full construction details and quality verification are in Suppl.~B.
 
\subsection{Alignment Strategies}
\label{sec:strategies}
 
We evaluate each task under three alignment strategies (Table~\ref{tab:strategies}). \textbf{Visual} (baseline) presents diagram images and video frames with no text mediation, exposing the full depiction gap. \textbf{Visual+Text} (V+T) appends a multi-dimension text description after each diagram image, providing a text bridge while retaining visual input. \textbf{Text Only} replaces diagram images entirely with text descriptions, eliminating the depiction gap.
 
\begin{table}[tb]
  \caption{Alignment strategies.}
  \label{tab:strategies}
  \small
  \begin{tabular}{llll}
    \toprule
    Strategy & Plan modality & Execution & Depiction gap? \\
    \midrule
    Visual & Diagram image & Video & Full gap \\
    Visual+Text & Diagram + text & Video & Gap + text bridge \\
    Text Only & Text description & Video & Eliminated \\
    \bottomrule
  \end{tabular}
\end{table}
 
Text Only eliminates the depiction gap by replacing diagrams with text, so it no longer tests cross-depiction alignment per se. It serves as a diagnostic control: comparing Text Only vs.\ Visual isolates the contribution of visual plan parsing, while comparing V+T vs.\ Visual reveals whether text provides information beyond what the visual encoder extracts.
 
To enable the text-mediated strategies, we generate 132 step descriptions from manual diagram images using Claude Opus 4.6~\cite{anthropic2026claude} in a structured multimodal prompt. Each diagram is independently described along 8 dimensions: parts, action, tools, spatial orientation, result state, warnings, fasteners, and arrow directions. Each description is then cross-validated against ground-truth part lists and connection annotations from the source dataset: 127 of 132 (96.2\%) are fully consistent with ground truth, and the 5 inconsistencies are traced to source data issues (blank images, duplicate files, annotation errors), not model failures. A post-hoc audit identified and removed cross-step references (e.g., ``the frame from step~2'') in 23 descriptions to prevent ordering leakage in the text-mediated evaluation settings. Full annotation details, prompt template, and cleaning procedure are in Suppl.~B.

\section{Experiments}
 
\subsection{Models}
 
We evaluate 17 open-source VLMs from 8 families spanning 2B to 38B parameters (Table~\ref{tab:models}), plus two proprietary models (Gemini 3.1 Pro, Gemini 3 Flash~\cite{geminiteam2025geminifamilyhighlycapable}). Our open-source selection covers all major families as of early 2026, includes 3 size tiers (small $\sim$2--4B, medium $\sim$7--12B, large $\sim$27--38B) for scaling analysis, and prioritizes widely deployed model families~\cite{Bai2025Qwen25VLTR}. Both Gemini models are evaluated under the Visual setting only via API. Implementation details (thinking-model handling, API configuration) are in Suppl.~C.
 
\begin{table}[tb]
  \caption{Evaluated model families.}
  \label{tab:models}
  \small
  \begin{tabular}{lll}
    \toprule
    Family & Sizes & Architecture \\
    \midrule
    Qwen2.5-VL & 3B, 7B & Dense, M-RoPE \\
    Qwen3-VL & 2B, 8B, 30B-A3B & Dense/MoE, DeepStack \\
    Qwen3.5 & 2B, 9B, 27B & Dense, GDN \\
    InternVL3.5 & 2B, 8B, 38B & Dense, InternViT \\
    Gemma3 & 4B, 12B, 27B & Dense \\
    GLM-4.1V & 9B & Dense, thinking \\
    LLaVA-OV & 8B & Dense, SigLIP \\
    MiniCPM-V & 8B & Dense \\
    \bottomrule
  \end{tabular}
\end{table}
 
\subsection{Evaluation Protocol}
 
All models are evaluated zero-shot with greedy decoding (max new tokens = 1024, bfloat16). We run 17 models $\times$ 3 strategies = 51 evaluation runs, each covering all 6 tasks. Prompts use interleaved image-text format with task-specific system context. Answer extraction uses multi-priority regex matching; the overall parse rate is 94.1\% across all runs, with failures counted as incorrect. Accuracy is the primary metric (4-way MC: chance = 25\%; binary: chance = 50\%). Full prompt templates and per-model parse rates are in Suppl.~A and~C.

\subsection{Three-Layer Mechanistic Analysis}
\label{sec:mechprotocol}

The benchmark results establish what fails but not why. To understand the mechanism behind cross-depiction failure, we probe the VLM processing pipeline at three successive levels. Each level tests a specific hypothesis about where the depiction gap originates and how text mediates the model's reasoning. We analyze four representative $\sim$8B-scale models: Qwen2.5-VL-7B, Qwen3-VL-8B, Qwen3.5-VL-9B, and InternVL3.5-8B. Layer~1 operates on 113 matched diagram--video step pairs extracted from the dataset; Layers~2 and~3 operate on 100 T1 benchmark questions under both Visual and V+T conditions. Full implementation details for all three layers are in Suppl.~D.

\subsubsection{Layer 1: Do Diagrams and Video Share a Representational Space?}
\label{sec:layer1protocol}

If the visual encoder maps diagrams and video frames into separate regions of its feature space, no downstream reasoning can recover alignment. We test this with three complementary metrics on frozen ViT features. For each of the 113 steps, we extract one mean-pooled ViT vector for the diagram image ($\mathbf{z}^{d} \in \mathbb{R}^{d}$) and one for the video segment ($\mathbf{z}^{v} \in \mathbb{R}^{d}$, averaged across all sampled frames of that step).

Centered Kernel Alignment (CKA)~\cite{Kornblith2019SimilarityON} measures whether the diagram representations and video representations, taken as two sets, share the same geometric structure. Concretely, we stack the 113 diagram vectors into $\mathbf{X} \in \mathbb{R}^{113 \times d}$ and the 113 video vectors into $\mathbf{Y} \in \mathbb{R}^{113 \times d}$, center both matrices, and compute:
\begin{equation}
  \text{CKA}(\mathbf{X}, \mathbf{Y}) = \frac{\| \mathbf{Y}^{\top} \mathbf{X} \|_F^2}{\| \mathbf{X}^{\top} \mathbf{X} \|_F \cdot \| \mathbf{Y}^{\top} \mathbf{Y} \|_F}
\end{equation}
CKA = 1 means the two sets are arranged identically in feature space; CKA = 0 means they occupy completely unrelated subspaces. We compute CKA at both the ViT output and the post-merger (projector output) level.

A linear probe~\cite{Alain2016UnderstandingIL} tests whether frozen video features encode enough temporal structure to tell whether two video frames depict the same assembly step. Given two video frames with ViT representations $\mathbf{r}_1, \mathbf{r}_2 \in \mathbb{R}^{d}$, we train a logistic regression classifier on the concatenated feature $[\mathbf{r}_1; \mathbf{r}_2; |\mathbf{r}_1 - \mathbf{r}_2|] \in \mathbb{R}^{3d}$ to predict whether the two frames come from the same step (chance = 50\%). Negative pairs are drawn from different steps of the same product so that they share most parts and differ only in configuration. If the probe performs at chance, the encoder does not distinguish assembly steps even within a single depiction.

Cross-modal retrieval directly tests diagram--video alignment: for each of the 113 diagrams, we retrieve the nearest video frame from a gallery of 2,546 frames by cosine similarity and check whether it belongs to the matching step. We report Recall@1 and Recall@10.

\subsubsection{Layer 2: Does Text Reduce the Model's Use of Diagram Information?}
\label{sec:layer2protocol}

The benchmark shows that adding text helps instruction comprehension but hurts cross-depiction alignment. We hypothesize that text reduces the model's reliance on diagram representations when forming its prediction. To test this, we run each model on 100 T1 questions under both Visual and V+T conditions and inspect the LLM's internal representations following Li~\etal{}~\cite{Li2025TheHL}.

Specifically, we extract the final-layer hidden state at the last input token, $\mathbf{h}_{\text{last}} \in \mathbb{R}^{d}$. This is the single vector from which the model generates its answer. We also compute, for each input modality $m$, the average hidden state over all tokens belonging to that modality: $\bar{\mathbf{h}}^{m} \in \mathbb{R}^{d}$, where $m \in \{\text{diagram, video, text}\}$. The modality influence score measures how much the prediction vector resembles each modality's representation:
\begin{equation}
  s_m = \cos(\mathbf{h}_{\text{last}},\; \bar{\mathbf{h}}^{m})
\end{equation}
A higher $s_m$ means modality $m$ has a stronger presence in the prediction. If $s_{\text{diagram}}$ drops from Visual to V+T, the model is shifting away from diagram information when text is available.

\subsubsection{Layer 3: Does Text Divert Attention from Visual Inputs?}
\label{sec:layer3protocol}

Layer~2 measures the net effect on the prediction vector but does not reveal how information flows through the model. We complement it with an attention-level analysis~\cite{Golovanevsky2024WhatDV} that directly shows how the model allocates its processing across modalities.

For Qwen3-VL-8B on the same 100 T1 questions under both Visual and V+T, we extract attention weights at 6 evenly spaced decoder layers. At each layer $l$, we look at the attention distribution from the last input token (the position that produces $\mathbf{h}_{\text{last}}$) and sum the weights falling on tokens of each modality:
\begin{equation}
  a_m^{(l)} = \sum_{i \in \mathcal{I}_m} \mathrm{softmax}\!\left(\frac{\mathbf{q}_{\text{last}}^{(l)} \cdot \mathbf{K}^{(l)\top}}{\sqrt{d_k}}\right)_{\!i}
\end{equation}
where $\mathcal{I}_m$ is the set of token positions belonging to modality $m$. We report $a_m$ averaged across all attention heads and the 6 probed layers. If adding text causes $a_{\text{diagram}}$ and $a_{\text{video}}$ to decrease while $a_{\text{text}}$ increases, this confirms that text diverts the model's processing away from visual inputs.

\begin{table*}[tb]
  \caption{Main results under the Visual (baseline) setting. Best open-source per column in \textbf{bold} (top), proprietary in \textbf{bold} (bottom). Full results (V+T, Text Only) in Suppl.~C.}
  \label{tab:mainresults}
  \small
  \begin{tabular}{ll r rrrrrr}
    \toprule
    Family & Model & Size & T1 Match & T2 Verify & T3 Locate & T4 Predict & D1 Exec. & D2 Plan \\
    \midrule
    Qwen2.5-VL & Qwen2.5-VL-3B & 3B & 42.8 & 51.1 & 35.6 & 28.9 & 48.3 & 52.3 \\
    & Qwen2.5-VL-7B & 7B & 49.1 & 50.9 & 35.0 & 36.8 & 46.0 & 53.8 \\
    Qwen3-VL & Qwen3-VL-2B & 2B & 42.2 & 50.0 & 29.6 & 34.8 & 50.0 & 26.2 \\
    & Qwen3-VL-8B & 8B & 53.1 & 56.6 & 49.4 & 39.7 & 58.3 & 58.5 \\
    & Qwen3-VL-30B-A3B & 30B & 48.8 & 58.3 & 50.6 & 34.3 & 60.0 & 56.9 \\
    Qwen3.5 & Qwen3.5-2B & 2B & 44.4 & 56.6 & 32.9 & 36.3 & 51.4 & 36.9 \\
    & Qwen3.5-9B & 9B & 57.8 & \textbf{63.7} & 46.7 & 38.2 & 63.1 & 58.5 \\
    & Qwen3.5-27B & 27B & \textbf{59.4} & 62.9 & \textbf{59.3} & \textbf{41.2} & \textbf{63.7} & \textbf{70.8} \\
    InternVL3.5 & InternVL3.5-2B & 2B & 33.4 & 50.3 & 29.9 & 23.0 & 48.6 & 20.0 \\
    & InternVL3.5-8B & 8B & 39.4 & 53.7 & 36.5 & 31.4 & 49.4 & 50.8 \\
    & InternVL3.5-38B & 38B & 54.4 & 61.4 & 47.3 & 37.7 & 61.4 & 67.7 \\
    Gemma3 & Gemma3-4B & 4B & 39.4 & 50.3 & 27.8 & 29.4 & 47.7 & 20.0 \\
    & Gemma3-12B & 12B & 35.3 & 49.7 & 35.9 & 28.4 & 49.1 & 32.3 \\
    & Gemma3-27B & 27B & 43.1 & 55.7 & 37.1 & 31.4 & 53.7 & 41.5 \\
    GLM-4.1V & GLM-4.1V-9B & 9B & 48.4 & 55.7 & 43.7 & 35.8 & 50.3 & 47.7 \\
    LLaVA-OV & LLaVA-OV-8B & 8B & 35.3 & 46.3 & 27.8 & 29.4 & 41.4 & 27.7 \\
    MiniCPM-V & MiniCPM-V-4.5 & 8B & 49.7 & 55.7 & 41.0 & 32.8 & 50.0 & 50.8 \\
    \midrule
    \multicolumn{8}{l}{\textit{Proprietary}} \\
    Gemini 3.1 & Gemini-3.1-Pro & --- & 62.8 & 65.1 & 65.0 & 41.7 & 67.4 & 76.9 \\
    Gemini 3 & Gemini-3-Flash & --- & \textbf{65.3} & \textbf{68.6} & \textbf{65.6} & \textbf{43.1} & \textbf{71.1} & \textbf{81.5} \\
    \midrule
    --- & \textit{Random} & --- & 25.0 & 50.0 & 25.0 & 25.0 & 50.0 & 25.0 \\
    \bottomrule
  \end{tabular}
\end{table*}

\section{Results and Analysis}
 
Table~\ref{tab:mainresults} presents the main results under the Visual (baseline) setting. Full results across all three strategies are in Suppl.~C. We first report behavioral findings from the benchmark evaluation (\S\ref{sec:benchresults}), then present the three-layer mechanistic analysis that explains these findings (\S\ref{sec:mechresults}).
 
\subsection{Benchmark Results}
\label{sec:benchresults}
 
\paragraph{Cross-depiction alignment accuracy is low, and drops further when procedural reasoning is required.}
 
T1 accuracy (4-way MC, chance = 25\%) ranges from 33.4\% (InternVL3.5-2B) to 59.4\% (Qwen3.5-27B) among open-source models, averaging 45.6\%. The two proprietary models reach 62.8\% (Gemini 3.1 Pro) and 65.3\% (Gemini 3 Flash), improving over the best open-source model by only 3--6\pp{}. Even the strongest model plateaus well below deployment-grade reliability. Performance drops further when procedural reasoning is required: the average T1$\to$T4 drop is 12.2\pp{} (45.6\%$\to$33.5\%), meaning that predicting the next step is substantially harder than recognizing the current step. Both proprietary models show similar drops ($-$21.1\pp{} and $-$22.2\pp{}), indicating that larger-scale pre-training does not alleviate this difficulty.
 
\begin{figure}[tb]
  \centering
  \includegraphics[width=0.85\columnwidth]{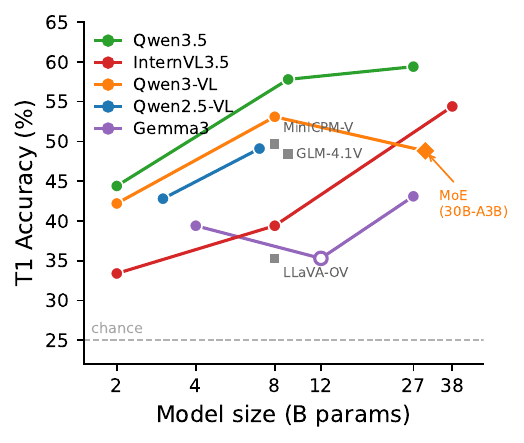}
  \caption{T1 accuracy vs.\ model size (log scale). Different families scale at different rates, and generational improvements at $\sim$8B consistently exceed within-family scaling.}
  \Description{Line chart of T1 accuracy vs model size showing non-parallel scaling curves across families.}
  \label{fig:scaling}
\end{figure}
 
\paragraph{Architecture family predicts cross-depiction alignment more reliably than parameter count.}
 
Figure~\ref{fig:scaling} shows that different model families scale at different rates: some families gain substantially with size while others plateau or even regress. At similar parameter counts ($\sim$8B), each Qwen generation yields larger gains than tripling parameters within a generation: Qwen2.5-VL-7B (49.1\%) $\to$ Qwen3-VL-8B (53.1\%) $\to$ Qwen3.5-9B (57.8\%) on T1. We cannot isolate whether architecture, training data, or training procedure drives the improvement, but the pattern is clear: upgrading to a newer model family at the same size is more effective than scaling up within a single family.
 
This pattern extends beyond the Qwen family. Qwen3-VL-8B (dense, 8B active, T1 = 53.1\%) outperforms Qwen3-VL-30B-A3B (MoE, 3B active, T1 = 48.8\%). Although the MoE variant has a larger total parameter count, its per-token active computation is less than half that of the dense model, and dense processing appears to be more effective for cross-depiction alignment. Gemma3 exhibits non-monotonic scaling: T1 follows 4B (39.4\%) $\to$ 12B (35.3\%$\downarrow$) $\to$ 27B (43.1\%), where the 12B model performs worse than 4B.
 
Both proprietary Gemini models improve over all open-source models on every task, but the gains are modest: +3--6\pp{} on T1, +4--7\pp{} on D1, and +6--11\pp{} on D2. On T4, the improvement is negligible (+0.5--1.9\pp{}). Notably, Gemini 3 Flash slightly outperforms the larger Gemini 3.1 Pro across all tasks. This directly addresses a natural objection: the depiction gap might simply reflect insufficient training data in open-source models. Proprietary models narrow the gap only marginally, confirming a structural rather than data-scale bottleneck. Developers building assembly guidance systems should evaluate newer model families (\eg{}, Qwen3.5) rather than solely scaling within the current family.
 
\begin{figure}[t]
  \centering
  \includegraphics[width=0.85\columnwidth]{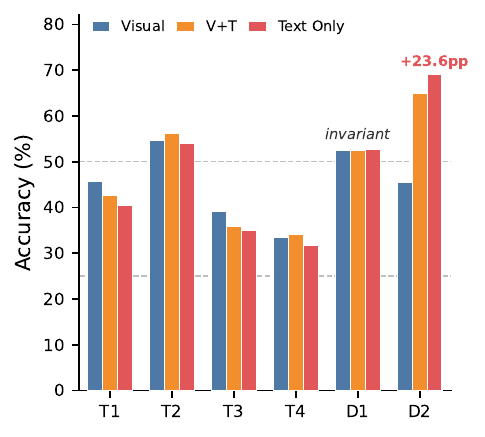}
  \caption{Average accuracy across 17 models per task and strategy. D1 is invariant to strategy (hard video ceiling), while D2 shows massive recovery with text (+23.6\pp{}).}
  \Description{Grouped bar chart showing 6 tasks under 3 strategies. D1 bars are nearly identical; D2 bars increase sharply from Visual to Text Only.}
  \label{fig:bottleneck}
\end{figure}
 
\paragraph{Video understanding is a hard ceiling; assembly instruction understanding is recoverable via text.}

The two diagnostic tasks reveal contrasting bottleneck profiles (Figure~\ref{fig:bottleneck}). D1 (video discrimination, binary, chance = 50\%) shows that 11 of 17 open-source models score below 53\%. Even the best proprietary model reaches only 71.1\%, and the best open-source model only 63.7\%. Video understanding is consistently the weakest link across all models tested.

D2 (assembly instruction ordering, 4-way MC, chance = 25\%, $N$ = 65) presents a different picture. Visual scores range from 20.0\% to 70.8\%; replacing diagrams with text yields an average gain of +23.6\pp{}. All 17 open-source models score higher on Text Only than Visual for D2, confirming that visual instruction understanding consistently lags text-based reasoning. The severity varies by architecture: LLaVA-OV shows the most extreme gap ($\Delta$ = +47.7\pp{}), while two models (Gemma3-4B, InternVL3.5-2B) score at or below the 25\% random baseline on D2 Visual (both 20.0\%), indicating that their visual encoders actively mislead on technical diagrams. D2's small sample size ($N$ = 65) limits per-model comparisons, but the aggregate trend across 17 models is robust.
 
\begin{table}[tb]
  \caption{Average strategy effect across 17 models.}
  \label{tab:avgeffect}
  \small
  \begin{tabular}{l rr l}
    \toprule
    Task & V$\to$V+T & V$\to$Text & Interpretation \\
    \midrule
    T1 (match) & $-$3.1 & $-$5.1 & Text hurts alignment \\
    T2 (verify) & +1.6 & $-$0.7 & Roughly neutral \\
    T3 (locate) & $-$3.2 & $-$4.2 & Text hurts alignment \\
    T4 (predict) & +0.5 & $-$1.7 & Roughly neutral \\
    D1 (video) & +0.1 & +0.1 & No effect (expected) \\
    D2 (instr.) & +19.5 & +23.6 & Text massively helps \\
    \bottomrule
  \end{tabular}
\end{table}
 
\paragraph{Text helps assembly instruction comprehension but does not improve cross-depiction alignment.}
 
The D2 results show that text can recover assembly instruction understanding. A natural question is whether this improvement transfers to cross-depiction alignment tasks. Table~\ref{tab:avgeffect} summarizes the average strategy effect across all 17 models. It does not transfer: 12 of 17 models show negative T1 delta under V+T, while all 17 show positive D2 delta. Text helps models understand assembly instructions but does not improve their ability to align instructions with video execution. Per-model strategy breakdowns are in Suppl.~C.
 
On T1, most models perform best under the Visual setting, with accuracy declining as text is added (V+T) or replaces diagrams entirely (Text Only). However, InternVL3.5-2B and InternVL3.5-8B show the reverse: Text Only $>$ Visual on T1 (37.5 vs.\ 33.4; 42.8 vs.\ 39.4). These models' visual encoders are so weak on diagrams that replacing diagrams with text slightly improves the instruction-understanding side. Yet the T1 gains are minimal (+3--4\pp{}), far smaller than the corresponding D2 gains (+25\pp{}). This confirms that video understanding, not instruction understanding, is the binding constraint on alignment: even when text substantially improves instruction comprehension, alignment barely improves because the model cannot adequately process the video side. This reversal disappears at scale (InternVL3.5-38B follows the majority pattern), suggesting it reflects visual encoder capacity at small model sizes.
 
The benchmark results above establish what fails and where: cross-depiction alignment accuracy is low (T1), drops further with procedural reasoning (T4), resists scaling (proprietary models), and is bottlenecked by video understanding (D1) rather than assembly instruction understanding (D2, recoverable via text). The mechanistic analysis below explains why.

\subsection{Mechanistic Analysis}
\label{sec:mechresults}
 
We now present results from the three-layer analysis described in \S\ref{sec:mechprotocol}, proceeding from visual encoding (Layer~1) through representation formation (Layer~2) to attention routing (Layer~3).

\begin{table}[tb]
  \caption{Layer~1: representation analysis across 4 models on 113 step pairs. CKA: diagram--video representational alignment (1 = identical, 0 = disjoint). Probe: video frame same/different step classification (chance = 50\%). R@$K$: diagram$\to$video retrieval at the merger level (gallery = 2,546 frames).}
  \label{tab:layer1}
  \small
  \begin{tabular}{l cc cc cc}
    \toprule
    & \multicolumn{2}{c}{CKA} & \multicolumn{2}{c}{Probe Acc} & \multicolumn{2}{c}{Retrieval} \\
    \cmidrule(lr){2-3} \cmidrule(lr){4-5} \cmidrule(lr){6-7}
    Model & ViT & Merger & ViT & Merger & R@1 & R@10 \\
    \midrule
    Qwen2.5-VL-7B & .006 & .137 & 48.9 & 61.3 & 5.3 & 15.0 \\
    Qwen3-VL-8B & .001 & .235 & 68.7 & 66.8 & 1.8 & 8.0 \\
    Qwen3.5-VL-9B & .006 & .105 & 52.8 & 61.5 & 4.4 & 12.4 \\
    InternVL3.5-8B & .101 & .075 & 52.7 & 68.3 & 0.0 & 8.8 \\
    \bottomrule
  \end{tabular}
\end{table}

\subsubsection{Layer 1: Diagrams and Video Occupy Disjoint Subspaces}
\label{sec:layer1results}

All three metrics consistently show that the visual encoder does not create a shared space for diagrams and video (Table~\ref{tab:layer1}). CKA at the ViT level is near zero for all four models (0.001--0.101), meaning diagrams and video frames have completely different representational geometries. The linear probe on video frame pairs performs at chance for 3 of 4 models at the ViT level (48.9\%--52.8\%), indicating that even within the video modality, frozen encoder features barely distinguish assembly steps. Cross-modal retrieval R@1 is near zero across all models, confirming that the ViT produces no shared space in which a diagram and its matching video segment are neighbors.

Post-merger (projector output) CKA improves but remains low (best: 0.235 for Qwen3-VL-8B), and probe accuracy rises to 61--68\%. The projector adds some task-relevant structure, but it is far from sufficient to bridge the depiction gap.

\begin{figure}[tb]
  \centering
  \includegraphics[width=0.85\columnwidth]{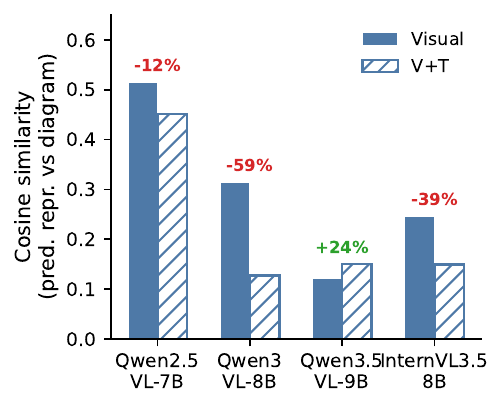}
  \caption{Diagram influence on the LLM prediction representation, measured by cosine similarity between the last-token hidden state and diagram token hidden states. Adding text (V+T) reduces diagram influence in 3 of 4 models ($n$=100 T1 questions per model).}
  \Description{Grouped bar chart of cosine similarity for 4 models under Visual vs V+T.}
  \label{fig:cosineshift}
\end{figure}

\subsubsection{Layer 2: Text Reduces the Model's Use of Diagram Information}
\label{sec:layer2results}

Diagram influence on the prediction representation drops in 3 of 4 models when text is added (Figure~\ref{fig:cosineshift}). The effect is strongest in Qwen3-VL-8B ($-$59\%) and clearly present in Qwen2.5-VL-7B ($-$12\%) and InternVL3.5-8B ($-$39\%). Qwen3.5-VL-9B is the exception ($+$24\%), which we attribute to its GDN hybrid attention architecture that distributes information uniformly across modalities, producing low cosine values under both conditions. This result directly explains why V+T degrades T1 performance: the model shifts toward text-based reasoning and away from the diagram features needed for visual spatial matching.

\begin{figure}[tb]
  \centering
  \includegraphics[width=0.85\columnwidth]{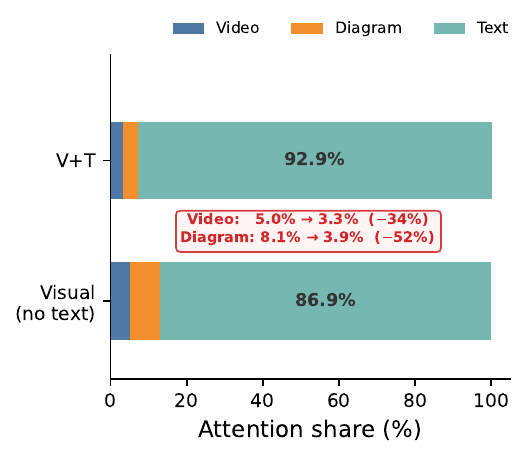}
  \caption{Per-modality attention share in Qwen3-VL-8B on T1 ($n$=100). Adding text halves diagram attention ($-$52\%) and reduces video attention ($-$34\%).}
  \Description{Horizontal stacked bars showing attention redistribution from diagrams and video toward text under V+T.}
  \label{fig:attnshift}
\end{figure}

\subsubsection{Layer 3: Text Absorbs Attention from Visual Modalities}
\label{sec:layer3results}

The attention analysis provides converging evidence for the distraction mechanism identified in Layer~2, at the level of information routing. Under the Visual condition, Qwen3-VL-8B allocates 8.1\% of last-token attention to diagram tokens and 5.0\% to video tokens (Figure~\ref{fig:attnshift}). Adding text halves diagram attention to 3.9\% and reduces video attention to 3.3\%, with text tokens absorbing both. This redistribution is consistent across all 6 probed decoder layers.

\section{Discussion}

\paragraph{Constructive insights.}
Our results yield several actionable insights for practitioners and researchers working on 2D-manual-based assembly guidance. First, no single alignment strategy dominates all sub-tasks: Visual input performs best for cross-depiction matching (T1, T3), while Text Only is preferable for assembly instruction comprehension (D2). A practical system should route sub-tasks to different strategies accordingly. Second, model selection should prioritize architecture family over parameter count. At similar scales, newer dense architectures (e.g., Qwen3.5) consistently outperform both older families and larger MoE variants, offering better cross-depiction alignment without additional compute cost. Third, our mechanistic analysis localizes the bottleneck to the visual encoder: diagrams and video occupy disjoint representational subspaces (CKA~$\approx$~0), and adding text actively distracts models from visual processing rather than supplementing it. This points to visual encoding as the primary target for future improvement. Concretely, fine-tuning on scaled schematic-to-real paired data, or training visual encoders with cross-depiction contrastive objectives, are promising directions for closing the gap at its root. 
 
\paragraph{Limitations and future work.}
Our benchmark covers IKEA furniture assembly only; whether the findings generalize to other assembly domains is untested. The benchmark tests foundational alignment capabilities, higher-level guidance functions such as error detection would be explored in the future. Furthermore, broader proprietary coverage would strengthen generality for the future work.

\section{Conclusion}

We present \bench{}, the first cross-depiction alignment benchmark for 2D-manual-based assembly guidance, evaluating 19 VLMs across 1,623 questions under three alignment strategies with a three-layer mechanistic analysis. Our findings show that assembly instruction understanding is recoverable via text, but this gain does not transfer to cross-depiction alignment. Architecture family predicts performance more reliably than parameter count, and even proprietary models improve by only 3--6\pp{}. Video understanding remains a hard bottleneck unaffected by any strategy. Mechanistically, diagrams and video occupy disjoint ViT subspaces, and adding text actively redirects both hidden-state reliance and attention away from visual modalities. These results motivate cross-depiction contrastive pre-training and adapter tuning on schematic-to-real paired data as direct paths to closing the gap.


\bibliographystyle{ACM-Reference-Format}
\bibliography{references}

\appendix

\section{Prompt Templates}
\label{app:prompts}

All tasks share the same structure: a system context, interleaved image-text content, a question with answer options, and a shared answer format instruction: ``Answer with the letter first, then briefly explain. Format: `Answer: X. Reason: ...'\,''

\subsection{Strategy Variations}

The three alignment strategies modify prompts as follows.

\textbf{Visual (baseline).} Diagram images are presented directly. System context references ``INSTRUCTION DIAGRAMS'' and ``wordless assembly instruction diagrams.''

\textbf{Visual+Text (V+T).} Same as Visual, but each diagram image is followed by a text annotation: ``(Description of the diagram above: \{text\})''. The system context adds: ``Each instruction diagram is accompanied by a text description that describes the content of that diagram. The text description is a supplementary annotation of the diagram.''

\textbf{Text Only.} All diagram images are replaced with text descriptions. System context changes ``INSTRUCTION DIAGRAMS'' to ``INSTRUCTION DESCRIPTIONS'' and ``wordless assembly instruction diagrams'' to ``text descriptions of assembly steps from a furniture manual.''

\subsection{Full System Contexts (Visual Setting)}

Below we show the system context and content structure for each task type under the Visual (baseline) setting.

\begin{promptbox}{T1: Step Recognition}
\textbf{System Prompt:}
You are evaluating a furniture assembly task. You will see:
1.~VIDEO FRAMES: Sequential frames captured from a real-world assembly video, showing a person physically assembling furniture.
2.~INSTRUCTION DIAGRAMS: Wordless assembly instruction diagrams (like IKEA manuals) that illustrate individual assembly steps using schematic drawings.
Your task is to determine which instruction diagram corresponds to the action shown in the video frames. Compare the physical action in the video against each diagram carefully.
\tcblower
\textbf{Content:} Video frames (N frames): [4--8 frame images] $\to$ Question $\to$ A) [diagram] B) [diagram] C) [diagram] D) [diagram] $\to$ answer instruction.
\end{promptbox}

\begin{promptbox}{T2: Action Verification}
\textbf{System Prompt:}
You are evaluating a furniture assembly task. You will see:
1.~VIDEO FRAMES: Sequential frames from a real-world assembly video.
2.~INSTRUCTION DIAGRAM: A single wordless assembly instruction diagram.
Your task is to judge whether the action shown in the video frames matches the assembly step depicted in the instruction diagram.
\tcblower
\textbf{Content:} Video frames (N frames): [frame images] $\to$ Instruction diagram: [one diagram] $\to$ Question $\to$ A) Yes / B) No $\to$ answer instruction.
\end{promptbox}

\begin{promptbox}{T3: Progress Tracking}
\textbf{System Prompt:}
You are evaluating a furniture assembly task. You will see:
1.~VIDEO FRAMES: Sequential frames from a real-world assembly video showing the current state of assembly.
2.~ALL INSTRUCTION DIAGRAMS: The complete set of wordless instruction diagrams for this product, shown in order (Step 1, Step 2, \ldots).
Your task is to determine which step in the instruction sequence corresponds to what is currently happening in the video.
\tcblower
\textbf{Content:} Video frames (N frames): [frame images] $\to$ Complete instruction sequence: Step~1: [diagram] Step~2: [diagram] \ldots $\to$ Question $\to$ 4-way MC with text labels $\to$ answer instruction.
\end{promptbox}

\begin{promptbox}{T4: Next-Step Prediction}
\textbf{System Prompt:}
You are evaluating a furniture assembly task. You will see:
1.~VIDEO FRAMES: Sequential frames from a real-world assembly video showing the action currently being performed.
2.~INSTRUCTION DIAGRAMS: Candidate diagrams for the NEXT assembly step.
Your task is to identify which instruction diagram shows the step that comes AFTER the action shown in the video. You need to understand what is currently being done and reason about what logically follows.
\tcblower
\textbf{Content:} Current step (video): [frame images] $\to$ What comes next? $\to$ A) [diagram] B) [diagram] C) [diagram] D) [diagram] $\to$ answer instruction.
\end{promptbox}

\begin{promptbox}{D1: Cross-View Matching}
\textbf{System Prompt:}
You are evaluating a furniture assembly task. You will see two sets of video frames (Clip~A and Clip~B) from different recordings of the same furniture product being assembled. Your task is to determine whether both clips show the same assembly step or different steps.
\tcblower
\textbf{Content:} Clip~A (N frames): [images] $\to$ Clip~B (N frames): [images] $\to$ Question $\to$ A) Same / B) Different $\to$ answer instruction. No diagrams; invariant across alignment strategies.
\end{promptbox}

\begin{promptbox}{D2: Sequence Ordering}
\textbf{System Prompt:}
You are evaluating a furniture assembly task. You will see three instruction diagrams (wordless assembly manual images) displayed in a shuffled (random) order. Your task is to determine the correct chronological assembly order of these diagrams by analyzing the assembly progression shown in each.
\tcblower
\textbf{Content:} Shuffled instruction diagrams: Image~1: [diagram] Image~2: [diagram] Image~3: [diagram] $\to$ Question $\to$ 4 permutation options $\to$ answer instruction. No video; plan-only task.
\end{promptbox}

\section{Data and Construction Details}
\label{app:data}

\begin{table}[htbp]
  \caption{Source dataset statistics (IKEA Manuals at Work).}
  \label{tab:datastats}
  \small
  \begin{tabular}{lr}
    \toprule
    Asset & Count \\
    \midrule
    Furniture products (6 categories) & 36 (29 used) \\
    Assembly videos & 98 \\
    Manual step diagrams (wordless PNG) & 133 \\
    Text descriptions (LLM-generated) & 132 \\
    Video frames (extracted) & 2,569 \\
    Benchmark questions & 1,623 \\
    \bottomrule
  \end{tabular}
\end{table}

\subsection{Excluded Products}

Of the 36 products in IKEA Manuals at Work, 7 are excluded:
\begin{itemize}
  \item \textbf{Too few steps ($\leq$2):} 4 products. T3 requires $\geq$3 steps for meaningful multiple-choice options. T4 and D2 require $\geq$3 consecutive steps for next-step prediction and ordering tasks.
  \item \textbf{Data quality issues:} 3 products. \textit{pahl}: 3 blank manual step images in the source dataset. \textit{sundvik}: step 3 reuses the same diagram image as step 2. \textit{grubban}: ground-truth annotation inconsistency at step 3 (annotated step does not match visible assembly action).
\end{itemize}

\subsection{Text Description Generation and Quality}

\paragraph{Annotation model and protocol.}
All 132 step descriptions are generated by Claude Opus 4.6~\cite{anthropic2026claude} using multimodal prompting. Each manual diagram image is processed independently (no cross-step context) with a structured prompt requesting descriptions along 8 dimensions:
\begin{enumerate}
  \item Parts: shape, count, material of individual pieces
  \item Action: assembly operation (attaching, screwing, inserting, etc.)
  \item Tools: visible or implied tools
  \item Spatial: orientation, insertion direction
  \item Result: post-step assembly state
  \item Warnings: X marks, caution symbols
  \item Fasteners: screw/bolt/dowel counts with part numbers
  \item Arrow directions: what arrows in the diagram indicate
\end{enumerate}
The average description length is 952 characters (range: 505--1,453). Descriptions are processed in 6 parallel batches with a total of $\sim$296K tokens ($\sim$2.2K per image).

\paragraph{Cross-validation against ground truth.}
Each description is compared against the source dataset's ground-truth part list (part IDs and counts) and connection annotations (which parts are joined). A description is flagged as inconsistent if the described parts or connections conflict with the ground truth. Of 132 descriptions, 127 (96.2\%) are fully consistent. The 5 flagged cases are all traced to source data issues rather than annotation errors: \textit{pahl} steps 0--2 have blank/corrupted PNG files in the source dataset, \textit{sundvik} step 3 reuses the step 2 image file, and \textit{grubban} step 3 has an inconsistency in the source ground-truth parts list.

\paragraph{Ordering leakage cleaning.}
The original manual images contain printed step numbers (e.g., ``1'', ``5''), and Claude occasionally produced cross-step references (e.g., ``the frame assembled in step~2''). In the text-mediated evaluation settings, such references would leak ordering information. A post-hoc audit found 23/132 descriptions (17\%) containing ``step~$N$'' cross-references, concentrated in early steps where Claude naturally referenced prior assembly. These were cleaned via regex replacement: ``from step~$N$'' $\to$ ``from a previous step'', ``same \ldots\ as step~$N$'' $\to$ ``similar to another step'', etc. (27 field modifications across 23 descriptions). A final audit confirmed 0 remaining cross-step references.

\subsection{Question Construction Pipeline}

All questions are deterministically constructed from ground-truth temporal alignment annotations provided by IKEA Manuals at Work~\cite{Liu2024IKEAMA}. Construction is fully reproducible.

\paragraph{Video frame sampling.}
For each assembly step, the source dataset provides a temporal segment [start, end] in each video. We uniformly sample 4--8 frames from this segment, with the number of frames scaling with segment duration. Transition zones at segment boundaries (first and last 10\% of the segment) are excluded to avoid ambiguous frames.

\paragraph{T1: Step Recognition.}
For each (product, step, video) tuple, we construct a 4-way MC question. The correct option is the diagram for the target step. The 3 distractors are diagrams from adjacent steps of the same product, ensuring that distractors share most parts and differ only in configuration (preventing coarse part-presence shortcuts). When fewer than 3 adjacent steps are available, we sample from the nearest available steps.

\paragraph{T2: Action Verification.}
For each (product, step, video) tuple, we construct two binary questions: one positive (matching diagram) and one negative (adjacent-step diagram). The positive/negative ratio across the full benchmark is 53\%/47\%. Option order (A=Yes/B=No vs. A=No/B=Yes) is randomized to prevent position bias.

\paragraph{T3: Progress Tracking.}
For each (product, step, video) tuple, we present all manual step diagrams for that product in order and ask which step is being performed. We construct 4-way MC by selecting the correct step plus 3 randomly sampled wrong steps from the same product.

\paragraph{T4: Next-Step Prediction.}
Similar to T1, but the correct answer is the diagram for the step \emph{after} the one shown in the video. This task requires both cross-depiction matching and procedural reasoning. Only steps with a valid next step are included (excluding the last step of each product).

\paragraph{D1: Cross-View Matching.}
For each product, we construct pairs of video frame sets. Positive pairs use frames from the same assembly step (possibly from different videos). Negative pairs use frames from different steps of the same product. The positive/negative ratio is 47\%/53\%. Option order is randomized.

\paragraph{D2: Sequence Ordering.}
For each product, we enumerate all positions where 3 consecutive manual step diagrams exist. The 3 diagrams are presented in shuffled order. We generate 4 ordering options: the correct chronological order plus 3 distractor permutations. Only consecutive triplets are used because non-adjacent steps are visually too dissimilar, allowing ordering through coarse part matching rather than genuine procedural reasoning. This constraint yields exactly 65 valid positions across all 29 products.

\subsection{Per-Product Statistics}

Table~\ref{tab:perproduct} reports the question count per task type for each of the 29 products.

\begin{table*}[htbp]
  \caption{Per-product benchmark statistics. Products sorted by category.}
  \label{tab:perproduct}
  \small
  \centering
  \begin{tabular}{llrrrrrrr}
    \toprule
    Product & Category & T1 & T2 & T3 & T4 & D1 & D2 & Total \\
    \midrule
    applaro\_3 & Bench & 3 & 10 & 3 & 0 & 3 & 1 & 20 \\
    tjusig & Bench & 20 & 16 & 22 & 12 & 21 & 3 & 94 \\
    applaro & Chair & 3 & 10 & 3 & 0 & 3 & 1 & 20 \\
    bekvam\_3017 & Chair & 30 & 16 & 30 & 24 & 23 & 3 & 126 \\
    grubban & Chair & 8 & 16 & 8 & 6 & 13 & 2 & 53 \\
    ingolf & Chair & 0 & 4 & 1 & 0 & 2 & 0 & 7 \\
    lisabo & Chair & 4 & 8 & 4 & 3 & 3 & 2 & 24 \\
    mammut\_1 & Chair & 0 & 16 & 0 & 0 & 23 & 0 & 39 \\
    marius & Chair & 0 & 16 & 0 & 0 & 22 & 0 & 38 \\
    nordviken & Chair & 3 & 6 & 3 & 3 & 2 & 2 & 19 \\
    poang\_1 & Chair & 6 & 12 & 6 & 0 & 11 & 1 & 36 \\
    poang\_2 & Chair & 34 & 16 & 34 & 30 & 21 & 4 & 139 \\
    ronninge & Chair & 4 & 8 & 4 & 4 & 3 & 5 & 28 \\
    stefan & Chair & 3 & 6 & 3 & 0 & 2 & 1 & 15 \\
    stig & Chair & 30 & 16 & 33 & 30 & 21 & 6 & 136 \\
    sundvik & Chair & 4 & 8 & 4 & 3 & 3 & 2 & 24 \\
    teodores & Chair & 28 & 16 & 34 & 22 & 21 & 4 & 125 \\
    vedbo & Chair & 6 & 12 & 6 & 0 & 11 & 1 & 36 \\
    pahl & Desk & 12 & 16 & 12 & 0 & 22 & 1 & 63 \\
    satsumas\_2 & Misc & 11 & 16 & 11 & 8 & 18 & 2 & 66 \\
    vesken & Misc & 20 & 14 & 20 & 15 & 21 & 2 & 92 \\
    vittsjo\_2 & Misc & 14 & 14 & 16 & 6 & 16 & 3 & 69 \\
    laiva & Shelf & 29 & 16 & 29 & 24 & 23 & 3 & 124 \\
    fjallbo & Table & 3 & 6 & 3 & 0 & 2 & 1 & 15 \\
    gladom & Table & 23 & 16 & 23 & 0 & 23 & 1 & 86 \\
    lunnarp & Table & 6 & 12 & 6 & 5 & 5 & 4 & 38 \\
    satsumas & Table & 3 & 6 & 3 & 0 & 2 & 1 & 15 \\
    tornviken & Table & 10 & 16 & 10 & 9 & 8 & 8 & 61 \\
    vittsjo\_1 & Table & 3 & 6 & 3 & 0 & 2 & 1 & 15 \\
    \midrule
    \textbf{Total} & & \textbf{320} & \textbf{350} & \textbf{334} & \textbf{204} & \textbf{350} & \textbf{65} & \textbf{1,623} \\
    \bottomrule
  \end{tabular}
\end{table*}

\section{Full Per-Model Results}
\label{app:fullresults}

Tables~\ref{tab:full_qwen}--\ref{tab:full_gemini} present the complete results for all 19 models (17 open-source + 2 proprietary) across alignment strategies and 6 task types. Open-source models are evaluated under all 3 strategies; proprietary Gemini models are evaluated under Visual only, as API-based evaluation does not permit the mechanistic analysis that motivates the multi-strategy design.

\begin{table}[htbp]
  \caption{Qwen family results (\%). V = Visual, V+T = Visual+Text, T = Text Only.}
  \label{tab:full_qwen}
  \small
  \begin{adjustbox}{max width=\columnwidth}
  \begin{tabular}{ll rrrrrr}
    \toprule
    Model & Set. & T1 & T2 & T3 & T4 & D1 & D2 \\
    \midrule
    Qwen2.5-VL-3B & V & 42.8 & 51.1 & 35.6 & 28.9 & 48.3 & 52.3 \\
    & V+T & 41.9 & 52.9 & 33.2 & 34.8 & 48.3 & 55.4 \\
    & T & 29.7 & 54.9 & 30.2 & 26.5 & 48.3 & 61.5 \\
    \addlinespace
    Qwen2.5-VL-7B & V & 49.1 & 50.9 & 35.0 & 36.8 & 46.0 & 53.8 \\
    & V+T & 42.2 & 50.3 & 34.1 & 33.8 & 46.0 & 67.7 \\
    & T & 39.4 & 50.6 & 32.0 & 29.9 & 46.0 & 72.3 \\
    \addlinespace
    Qwen3-VL-2B & V & 42.2 & 50.0 & 29.6 & 34.8 & 50.0 & 26.2 \\
    & V+T & 34.4 & 51.4 & 30.5 & 34.3 & 50.0 & 49.2 \\
    & T & 39.1 & 51.7 & 29.9 & 29.4 & 50.0 & 47.7 \\
    \addlinespace
    Qwen3-VL-8B & V & 53.1 & 56.6 & 49.4 & 39.7 & 58.3 & 58.5 \\
    & V+T & 45.0 & 59.1 & 45.8 & 36.8 & 58.3 & 80.0 \\
    & T & 43.1 & 55.7 & 37.7 & 30.4 & 58.3 & 83.1 \\
    \addlinespace
    Qwen3-VL-30B-A3B & V & 48.8 & 58.3 & 50.6 & 34.3 & 60.0 & 56.9 \\
    & V+T & 49.1 & 65.7 & 38.0 & 36.8 & 60.0 & 89.2 \\
    & T & 50.3 & 62.0 & 35.3 & 32.8 & 60.0 & 86.2 \\
    \addlinespace
    Qwen3.5-2B & V & 44.4 & 56.6 & 32.9 & 36.3 & 51.4 & 36.9 \\
    & V+T & 36.6 & 57.4 & 34.7 & 33.8 & 51.4 & 52.3 \\
    & T & 33.1 & 54.9 & 34.7 & 27.5 & 51.4 & 43.1 \\
    \addlinespace
    Qwen3.5-9B & V & 57.8 & 63.7 & 46.7 & 38.2 & 63.1 & 58.5 \\
    & V+T & 54.1 & 65.1 & 40.4 & 35.8 & 63.1 & 76.9 \\
    & T & 45.9 & 58.0 & 42.8 & 34.3 & 63.1 & 78.5 \\
    \addlinespace
    Qwen3.5-27B & V & 59.4 & 62.9 & 59.3 & 41.2 & 63.7 & 70.8 \\
    & V+T & 59.4 & 63.7 & 47.9 & 45.1 & 63.7 & 93.8 \\
    & T & 54.4 & 59.1 & 47.3 & 42.2 & 63.7 & 95.4 \\
    \bottomrule
  \end{tabular}
  \end{adjustbox}
\end{table}

\begin{table}[htbp]
  \caption{InternVL family results (\%).}
  \label{tab:full_internvl}
  \small
  \begin{adjustbox}{max width=\columnwidth}
  \begin{tabular}{ll rrrrrr}
    \toprule
    Model & Set. & T1 & T2 & T3 & T4 & D1 & D2 \\
    \midrule
    InternVL3.5-2B & V & 33.4 & 50.3 & 29.9 & 23.0 & 48.6 & 20.0 \\
    & V+T & 33.4 & 48.9 & 33.8 & 27.0 & 48.6 & 36.9 \\
    & T & 37.5 & 50.6 & 33.8 & 34.3 & 48.6 & 53.8 \\
    \addlinespace
    InternVL3.5-8B & V & 39.4 & 53.7 & 36.5 & 31.4 & 49.4 & 50.8 \\
    & V+T & 42.2 & 56.0 & 33.8 & 31.9 & 49.4 & 72.3 \\
    & T & 42.8 & 54.3 & 39.2 & 33.3 & 49.4 & 75.4 \\
    \addlinespace
    InternVL3.5-38B & V & 54.4 & 61.4 & 47.3 & 37.7 & 61.4 & 67.7 \\
    & V+T & 53.4 & 60.9 & 43.4 & 37.3 & 61.4 & 86.2 \\
    & T & 47.8 & 57.4 & 40.1 & 37.3 & 61.4 & 87.7 \\
    \bottomrule
  \end{tabular}
  \end{adjustbox}
\end{table}

\begin{table}[htbp]
  \caption{Gemma3 family results (\%).}
  \label{tab:full_gemma}
  \small
  \begin{adjustbox}{max width=\columnwidth}
  \begin{tabular}{ll rrrrrr}
    \toprule
    Model & Set. & T1 & T2 & T3 & T4 & D1 & D2 \\
    \midrule
    Gemma3-4B & V & 39.4 & 50.3 & 27.8 & 29.4 & 47.7 & 20.0 \\
    & V+T & 33.1 & 53.1 & 25.4 & 27.0 & 47.7 & 21.5 \\
    & T & 32.2 & 48.9 & 24.9 & 28.4 & 47.7 & 21.5 \\
    \addlinespace
    Gemma3-12B & V & 35.3 & 49.7 & 35.9 & 28.4 & 49.1 & 32.3 \\
    & V+T & 38.4 & 49.1 & 29.3 & 34.8 & 49.1 & 43.1 \\
    & T & 36.9 & 48.6 & 31.4 & 28.9 & 49.1 & 60.0 \\
    \addlinespace
    Gemma3-27B & V & 43.1 & 55.7 & 37.1 & 31.4 & 53.7 & 41.5 \\
    & V+T & 42.5 & 52.0 & 35.6 & 30.4 & 53.7 & 66.2 \\
    & T & 42.2 & 51.1 & 34.4 & 28.9 & 53.7 & 78.5 \\
    \bottomrule
  \end{tabular}
  \end{adjustbox}
\end{table}

\begin{table}[htbp]
  \caption{Other open-source model results (\%).}
  \label{tab:full_other}
  \small
  \begin{adjustbox}{max width=\columnwidth}
  \begin{tabular}{ll rrrrrr}
    \toprule
    Model & Set. & T1 & T2 & T3 & T4 & D1 & D2 \\
    \midrule
    GLM-4.1V-9B & V & 48.4 & 55.7 & 43.7 & 35.8 & 50.3 & 47.7 \\
    & V+T & 42.8 & 59.4 & 37.7 & 32.8 & 50.3 & 75.4 \\
    & T & 43.1 & 54.9 & 33.8 & 29.9 & 50.3 & 76.9 \\
    \addlinespace
    LLaVA-OV-8B & V & 35.3 & 46.3 & 27.8 & 29.4 & 41.4 & 27.7 \\
    & V+T & 35.0 & 52.9 & 31.1 & 32.4 & 41.4 & 60.0 \\
    & T & 30.9 & 52.0 & 29.0 & 30.9 & 41.4 & 75.4 \\
    \addlinespace
    MiniCPM-V-4.5 & V & 49.7 & 55.7 & 41.0 & 32.8 & 50.0 & 50.8 \\
    & V+T & 40.6 & 57.7 & 36.5 & 34.3 & 50.9 & 76.9 \\
    & T & 40.0 & 51.7 & 37.4 & 35.3 & 52.0 & 76.9 \\
    \bottomrule
  \end{tabular}
  \end{adjustbox}
\end{table}

\begin{table}[htbp]
  \caption{Proprietary model results (\%, Visual setting only). Gemini models are evaluated via API; multi-strategy and mechanistic analyses are not applicable.}
  \label{tab:full_gemini}
  \small
  \begin{adjustbox}{max width=\columnwidth}
  \begin{tabular}{ll rrrrrr}
    \toprule
    Model & Set. & T1 & T2 & T3 & T4 & D1 & D2 \\
    \midrule
    Gemini-3.1-Pro & V & 62.8 & 65.1 & 65.0 & 41.7 & 67.4 & 76.9 \\
    Gemini-3-Flash & V & 65.3 & 68.6 & 65.6 & 43.1 & 71.1 & 81.5 \\
    \bottomrule
  \end{tabular}
  \end{adjustbox}
\end{table}

\subsection{Per-Model Parse Rates}

The overall parse rate across all 51 evaluation runs (17 open-source models $\times$ 3 strategies) is 94.1\%. Answer extraction uses a multi-priority regex pipeline: (1)~match ``Answer: X'' pattern, (2)~match first word-boundary letter A--D, (3)~match single-letter response. For GLM-4.1V-9B, thinking blocks are stripped before extraction. Unparseable responses are counted as incorrect.

Most models achieve 100\% parse rates across all settings, including all Qwen models (8), all InternVL models (3), LLaVA-OV-8B, MiniCPM-V-4.5, and Gemma3-4B. The exceptions are: Gemma3-12B (98.0--99.9\%), Gemma3-27B (99.1--99.6\%), and GLM-4.1V-9B (95.0--98.1\%, due to thinking-mode output formatting). Both Gemini models achieve 99.6\% parse rate (1,616/1,623 parsed).

\subsection{Implementation Details}

All open-source models are evaluated in bfloat16 precision with greedy decoding (temperature=0, max new tokens=1024). Models are loaded with automatic device mapping on a single NVIDIA H200 (141\,GB). Attention implementation varies by architecture:
\begin{itemize}
  \item Flash Attention 2: Qwen2.5-VL, Qwen3-VL, Gemma3
  \item SDPA: Qwen3.5 (required by GDN hybrid attention)
  \item Trust remote code: InternVL, MiniCPM, GLM
\end{itemize}
Model weights are cached locally; no quantization is applied.

Both proprietary Gemini models (3.1 Pro, 3 Flash) are evaluated via the Google AI Studio API with temperature=0 and max output tokens=256, using the same prompts and answer extraction pipeline as open-source models.

\subsection{Strategy Effect on Representative Models}

Table~\ref{tab:suppl_strategyeffect} shows T1 and D2 accuracy across all 3 alignment strategies for 5 representative models spanning 4 families, 3 size tiers, and including reversed-preference cases (InternVL3.5-8B). This table complements the aggregate strategy effect analysis in the main paper.

\begin{table}[htbp]
  \caption{Strategy effect on 5 representative models.}
  \label{tab:suppl_strategyeffect}
  \small
  \begin{adjustbox}{max width=\columnwidth}
  \begin{tabular}{l rrr rrr}
    \toprule
    & \multicolumn{3}{c}{T1 Match} & \multicolumn{3}{c}{D2 Instr.} \\
    \cmidrule(lr){2-4} \cmidrule(lr){5-7}
    Model & V & V+T & Text & V & V+T & Text \\
    \midrule
    Qwen3-VL-8B & 53.1 & 45.0 & 43.1 & 58.5 & 80.0 & 83.1 \\
    Qwen3.5-27B & 59.4 & 59.4 & 54.4 & 70.8 & 93.8 & 95.4 \\
    InternVL3.5-8B & 39.4 & 42.2$\uparrow$ & 42.8$\uparrow$ & 50.8 & 72.3 & 75.4 \\
    LLaVA-OV-8B & 35.3 & 35.0 & 30.9 & 27.7 & 60.0 & 75.4 \\
    Gemma3-27B & 43.1 & 42.5 & 42.2 & 41.5 & 66.2 & 78.5 \\
    \bottomrule
  \end{tabular}
  \end{adjustbox}
  \\[2pt]
  {\footnotesize $\uparrow$ = reversed preference (Text Only $>$ Visual on T1).}
\end{table}

\subsection{Diagram Blindness Analysis}

\begin{figure}[htbp]
  \centering
  \includegraphics[width=0.85\columnwidth]{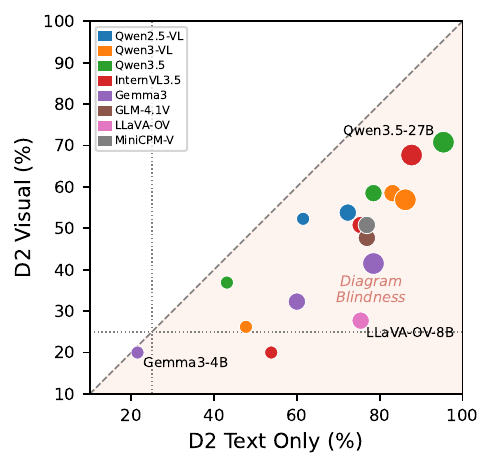}
  \caption{D2 Visual vs.\ Text Only accuracy for 17 VLMs. All models fall below the diagonal: visual assembly instruction understanding consistently lags text-based reasoning. Marker size indicates parameter count.}
  \Description{Scatter plot with all 17 models below the y=x diagonal. The shaded region below the diagonal is labeled Diagram Blindness.}
  \label{fig:suppl_diagramblind}
\end{figure}

Figure~\ref{fig:suppl_diagramblind} visualizes the D2 Visual vs.\ Text Only accuracy for all 17 open-source models. Every model falls below the $y=x$ diagonal, confirming that visual assembly instruction understanding consistently lags text-based reasoning across all architectures and scales.

\section{Mechanistic Analysis Details}
\label{app:mechdetails}

This section provides full implementation details for the three-layer mechanistic analysis described in Section~4.3 of the main paper.

\subsection{Feature Extraction (Layer 1)}

\paragraph{ViT representations.}
ViT last-layer representations are extracted via forward hook on the final transformer block output, then mean-pooled across all patch tokens per image to produce a single vector $\mathbf{z} \in \mathbb{R}^{d}$. For models with tile-based image processing (Qwen family), all tiles are flattened before averaging. For InternVL3.5, the CLS token of each tile is removed before averaging. All representations are stored in float32. ViT hidden dimensions vary by model: 1024 (InternVL3.5-8B), 1152 (Qwen3-VL-8B, Qwen3.5-VL-9B), 1280 (Qwen2.5-VL-7B).

\paragraph{Merger representations.}
Post-merger (projector output) representations are extracted similarly after the visual-to-LLM projection layer. Merger output dimensions match the LLM hidden dimension: 3584 (Qwen2.5-VL-7B), 4096 (Qwen3-VL-8B, Qwen3.5-VL-9B, InternVL3.5-8B).

\paragraph{CKA computation.}
For each of the 113 valid step pairs (steps with both a diagram and valid video segment), both the diagram and video representations are mean-pooled into single vectors. These are stacked across all 113 steps to form matrices $\mathbf{X} \in \mathbb{R}^{113 \times d}$ (diagrams) and $\mathbf{Y} \in \mathbb{R}^{113 \times d}$ (video). Both matrices are centered before computing linear CKA~\cite{Kornblith2019SimilarityON}:
$$\text{CKA}(\mathbf{X}, \mathbf{Y}) = \frac{\|\mathbf{Y}^{\top}\mathbf{X}\|_F^2}{\|\mathbf{X}^{\top}\mathbf{X}\|_F \cdot \|\mathbf{Y}^{\top}\mathbf{Y}\|_F}$$
CKA is dimension-invariant, enabling cross-architecture comparison. Bootstrap 95\% confidence intervals are computed from 1,000 resamples over the 113 step pairs.

\paragraph{Video temporal probe.}
Logistic regression (C=1.0, L-BFGS solver, max 1,000 iterations) with StandardScaler on concatenated features $[\mathbf{r}_1; \mathbf{r}_2; |\mathbf{r}_1 - \mathbf{r}_2|]$. Positive pairs: same step, different video frames. Negative pairs: different steps within the same product (hard negatives). Train/test split by product (every 5th product held out, deterministic seed 42). Maximum 20K training pairs and 4K test pairs. Positive:negative ratio is 1:4.

\paragraph{Cross-modal retrieval.}
For each of the 113 diagram representations, we compute cosine similarity against a gallery of all 2,546 individual video frame representations (L2-normalized). A retrieval is correct if any video frame from the matching step appears in the top $K$ results. Recall@$K$ reports the fraction of diagrams for which this holds. The main paper's Layer~1 table reports retrieval results at the post-merger level.

\subsection{LLM Hidden States (Layer 2)}

For each T1 question ($n$=100 per model), we run a forward pass under both Visual and V+T conditions and extract the LLM's final decoder layer hidden states via forward hook. The prediction representation $\mathbf{h}_{\text{last}} \in \mathbb{R}^{d}$ is the hidden state at the last input token (before generation begins). Per-modality representations $\bar{\mathbf{h}}^{m} \in \mathbb{R}^{d}$ are computed by mean-pooling hidden states over token positions belonging to each modality $m \in \{\text{video, diagram, text}\}$. The modality influence score is:
$$s_m = \cos(\mathbf{h}_{\text{last}},\; \bar{\mathbf{h}}^{m})$$
Token categorization is performed by scanning input token IDs for vision span markers (vision start/end tokens for Qwen; image context token ID for InternVL). Video spans appear first, followed by diagram spans. All remaining non-system tokens are classified as text. All cosine similarity computations use float32.

\subsection{Attention Analysis (Layer 3)}

\paragraph{Attention extraction.}
For each of 100 T1 questions under both Visual and V+T conditions, last-token attention is extracted from Qwen3-VL-8B. At each of 6 evenly spaced layers $l \in \{0, 7, 15, 23, 31, 35\}$ across the 36-layer decoder, we compute the attention distribution:
$$\mathbf{a}^{(l)} = \mathrm{softmax}\!\left(\frac{\mathbf{q}_{\text{last}}^{(l)} \cdot \mathbf{K}^{(l)\top}}{\sqrt{d_k}}\right)$$
from RoPE-rotated Q,K states. Attention weights are aggregated by modality ($a_m^{(l)} = \sum_{i \in \mathcal{I}_m} a_i^{(l)}$) and averaged across heads and layers. Grouped Query Attention (GQA) expansion: 8 KV heads are expanded to 32 Q heads via repeat-interleave. Softmax is computed in float32 for numerical stability. Reported attention shares are averaged across all 32 heads and all 6 probed layers. This requires $O(\mathrm{seq})$ memory per layer (not $O(\mathrm{seq}^2)$), enabling analysis on sequences up to 11K tokens without materializing full attention matrices.

\paragraph{Token categorization.}
Input tokens are categorized into modalities by scanning for vision span markers (vision start/end special tokens). Video spans appear first in the sequence, followed by diagram option spans. All remaining non-system, non-special tokens are classified as text/prompt tokens. Attention shares per modality are computed by summing the last-token attention weights over the corresponding token positions.

\end{document}